\documentclass[11pt,a4paper]{article}
\usepackage[hyperref]{acl2020}
\usepackage{times}
\usepackage{latexsym}
\usepackage{graphicx}
\usepackage{float}
\usepackage{array}
\usepackage{amssymb}
\usepackage{pifont}
\newcommand{\cmark}{\ding{51}}%

\usepackage{microtype}

\aclfinalcopy 

\newcommand\alexa{$^\diamondsuit$}
\newcommand\cmu{$^\spadesuit$}
\newcommand\aspace{\hspace{.75em}}

\title{DialoGLUE: A Natural Language Understanding Benchmark for Task-Oriented Dialogue}

\author{Shikib Mehri\thanks{\hspace{0.03in} This work was done while the first author was an intern at Amazon.}\cmu\aspace  Mihail Eric\alexa\aspace  Dilek Hakkani-Tur\alexa \\
\cmu Language Technologies Institute, Carnegie Mellon University \\
\alexa Amazon Alexa AI \\
\texttt{amehri@cs.cmu.edu, \{mihaeric, hakkanit\}@amazon.com}}

\date{}

\begin{document}
\maketitle
\begin{abstract}
A long-standing goal of task-oriented dialogue research is the ability to flexibly adapt dialogue models to new domains. To progress research in this direction, we introduce \textbf{DialoGLUE} (Dialogue Language Understanding Evaluation), a public benchmark consisting of 7 task-oriented dialogue datasets covering 4 distinct natural language understanding tasks, designed to encourage dialogue research in representation-based transfer, domain adaptation, and sample-efficient task learning. We release several strong baseline models, demonstrating performance improvements over a vanilla BERT architecture and state-of-the-art results on 5 out of 7 tasks, by pre-training on a large open-domain dialogue corpus and task-adaptive self-supervised training. Through the DialoGLUE benchmark, the baseline methods, and our evaluation scripts, we hope to facilitate progress towards the goal of developing more general task-oriented dialogue models.\footnote{\url{https://evalai.cloudcv.org/web/} \\
\url{challenges/challenge-page/708/}}
\end{abstract}

\section{Introduction}

One of the ultimate goals of task-oriented conversational systems is the ability to flexibly bootstrap new dialogue functionalities across diverse domains of user interest. For example, once we have successfully built a dialogue assistant that can handle restaurant booking queries, we would ideally like that assistant to quickly start servicing hotel reservation queries without too much additional manual effort. Unfortunately in the modern conversational assistant ecosystem, the work required to scale up functionalities is often linear with respect to the number of desired domains.

For modelling improvements to claim meaningful progress towards generality, the improvements must extend beyond a single dataset and instead hold across several different dialogue tasks and corpora. We argue that one of the core roadblocks in progressing the generality of conversational systems toward this desired state is a lack of standardization in both datasets and evaluation used by the community. These problems have been noted more broadly in the natural language understanding field, inspiring numerous efforts to propose unified benchmarks spanning multiple downstream tasks across different corpora with consolidated evaluation procedures~\cite{wang2018glue,Wang2019SuperGLUEAS,McCann2018TheNL}.   

While today there is a reasonable abundance of available corpora for building data-driven dialogue systems~\cite{Serban2018ASO}, little work has been done to bring together the efforts of researchers to reflect the properties we care about in our systems: statistical learning that is data-efficient and robustly transfers across domains and tasks.

In this work we propose DialoGLUE, a public benchmark consisting of 7 diverse task-oriented spoken-language datasets across 4 distinct natural language understanding tasks including intent prediction, slot-filling, semantic parsing, and dialogue state tracking. Many of these datasets include multiple system functionalities and in total, the DialoGLUE benchmark covers over 40 different domains. Our benchmark is designed to encourage dialogue research in representation-based transfer, domain adaptation, and sample-efficient task learning. Furthermore, it consists entirely of previously-published datasets that have reported results, as these resources have been vetted by the broader community to be sufficiently difficult and interesting.

As part of DialoGLUE, we also release evaluation scripts and competitive BERT-based baselines on the downstream tasks. We introduce \textsc{ConvBERT}, a BERT-base model which has been trained on a large open-domain dialogue corpus. In combination with task-adaptive pre-training and multi-tasking, \textsc{ConvBERT} matches or exceeds state-of-the-art results on five of the seven DialoGLUE datasets. Most notably, we attain a \textbf{+2.98} improvement in the joint goal accuracy over the best dialogue state tracking models on the MultiWOZ corpus. While our baselines demonstrate the efficacy of task-adaptive finetuning in transferring across datasets, there is still plenty of headroom in the aggregated benchmark scores encouraging further research.

In summary, the contributions of this work are four-fold: (1) a challenging task-oriented dialogue benchmark consisting of 7 distinct datasets across 4 domains, (2) standardized evaluation measures for reporting results (3) competitive baselines across the tasks, and  (4) a public leaderboard for tracking scientific progress on the benchmark.

The DialoGLUE leaderboard and evaluation scripts are hosted using the EvalAI platform \citep{EvalAI}. Our code and baseline models are open-sourced.\footnote{\url{https://github.com/alexa/DialoGLUE/}}

\section{Related Work}

\subsection{NLP Benchmarks}

In part, the development of unified benchmarks have helped drive progress towards more general models of language. The GLUE \citep{wang2018glue} and SuperGLUE benchmarks \citep{Wang2019SuperGLUEAS} have provided a consistent benchmark that allows pre-trained language models \citep{devlin2018bert,radford2018improving} to be evaluated on a variety of tasks.

While GLUE and SuperGLUE are concentrated on language understanding tasks, decaNLP \citep{McCann2018TheNL} consists of a broader set of NLP tasks including question answering, machine translation and summarization. 

Within conversational systems, the recently proposed Dialogue Dodecathlon is a benchmark for knowledge-grounded, situated, and multi-modal dialogue generation consisting of several open-domain dialogue datasets~\cite{shuster2019dialogue}. In contrast to the Dialogue Dodecathlon, DialoGLUE focuses on task-oriented dialogue and language understanding.

DialoGLUE is inspired by GLUE and SuperGLUE, however our efforts differ in that we aim to produce a benchmark for natural language understanding in the context of task-oriented dialogue. Task-oriented dialogue poses a unique set of challenges; dialogue is intrinsically goal-driven, multi-turn and often informal/noisy \citep{henderson2019convert,zhang2019dialogpt}. Through DialoGLUE, we hope to provide a benchmark for assessing models that aim to tackle these unique challenges across several tasks within task-oriented dialogue.

\subsection{Pre-trained Models}

Large-scale pre-training has attained significant performance gains across many tasks within NLP \citep{devlin2018bert,radford2018improving}. Through self-supervised pre-training on large natural language corpora, these models gain generalized language understanding capabilities that transfer effectively to downstream tasks \citep{wang2018glue}, including intent prediction \citep{chen2019bert,castellucci2019multi} and dialogue state tracking \citep{heck2020trippy}. However, recent work has demonstrated that leveraging dialog-specific pre-trained models, such as ConveRT \citep{henderson2019convert,casanueva2020efficient} obtains better results.

Large-scale pre-training on open-domain dialogue has demonstrated surprising generality, with models like DialoGPT \citep{zhang2019dialogpt}, Meena \citep{adiwardana2020towards} and Blender \citep{roller2020recipes} achieving response generation performance competitive with humans in certain settings. ConveRT has demonstrated that pre-training on open-domain dialogue transfers to task-oriented dialogue, with significant performance improvements over BERT on both intent prediction \citep{casanueva2020efficient} and slot filling \citep{coope2020span}. Inspired by these results, we fine-tune BERT \citep{devlin2018bert} on a large open-domain dialogue corpus prior to fine-tuning on the downstream tasks of DialoGLUE.

\subsection{Task-Adaptive Training}

Though large-scale pre-training results in strong performance when transferring to downstream tasks, performing self-supervised training on a target dataset allows the model to better adapt to the dataset prior to fine-tuning \citep{mehri2019pretraining,gururangan2020don}. Since our target datasets differ from the pre-training data, we hypothesize that task-adaptive training will allow pre-trained models to better adapt to task-oriented dialogue. 

\section{DialoGLUE Tasks}

The DialoGLUE benchmark consists of 7 different datasets spanning 4 different tasks: intent prediction, slot filling, semantic parsing and dialogue state tracking. These datasets all share the common goal of language understanding in the context of dialogue, making them suitable for DialoGLUE. 

\subsection{Intent Prediction}

Intent prediction is the task of extracting meaning from a natural language utterance in order to understand the user's goal \citep{hemphill-etal-1990-atis,coucke2018snips}. We leverage three different datasets for the task of intent prediction, all of which span several domains and consist of many different intents.  

\textbf{\textsc{banking77}} \citep{casanueva2020efficient} contains 13,083 utterances related to banking with 77 different fine-grained intents. Despite only consisting of a single domain, this dataset is challenging as it requires fine-grained differentiation betweeen very similar intents like \texttt{card payment wrong exchange rate} and \texttt{wrong exchange rate for cash withdrawal}.

\textbf{\textsc{clinc150}} \citep{larson-etal-2019-evaluation} contains 23,700 utterances spanning 10 domains (e.g., travel, kitchen/dining, utility, small talk, etc.) and 150 different intent classes. This dataset also consists of \textit{out of scope} utterances, which do not belong to any of the other intents and must therefore be classified as out of scope.

\textbf{\textsc{hwu64}} \citep{liu2019benchmarking} includes 25,716 utterances for 64 intents spanning 21 domains (e.g., alarm, music, IoT, news, calendar, etc.). The domains and intents of this dataset are similar to ones we expect users to ask a virtual assistant (e.g., Alexa, Google Assistant, Siri).

\citet{casanueva2020efficient} forego a validation set when using these datasets for intent prediction and instead only use a training and testing set. We instead designate a portion of the training set to be the validation set.

\subsection{Slot Filling}

Slot filling, a vital component of task-oriented dialogue systems, is the task of identifying values for pre-defined attributes in a natural language utterance. Following the set up of \citet{coope2020span}, we include two slot filling datasets in DialoGLUE.

\textbf{\textsc{restaurant8k}} \citep{coope2020span} comprises of 8,198 utterances from a commercial restaurant booking system and consists of 5 slots (date, time, people, first name, last name).

\textbf{\textsc{DSTC8}} \citep{rastogi2019towards} consists of slot annotations extracted by \citet{coope2020span} spanning 4 domains (buses, events, homes, rental cars) for a total of 5,569 utterances.

For both these datasets, the value for a particular slot will always be a contiguous span of the utterance. For some utterances, the expected slot (e.g., ``people'', ``time'') is provided. This allows an otherwise ambiguous utterance like \textit{``four''} to be interpreted either as \textit{``four people''} or \textit{``four o'clock''}.

\subsection{Semantic Parsing}

\textbf{\textsc{TOP}} \citep{gupta2018semantic} is a dataset of 44k utterances wherein each utterance is annotated with a hierarchical semantic representation. Each utterance has a top-level intent, which serves as the root of the tree. Every word of the utterance is a leaf node of the tree. The sub-trees correspond to different slots and intents, with the dataset having an average tree depth of 2.54. \textsc{TOP} covers two different domains, navigation (directions, distance, traffic) and events.

\subsection{Dialogue State Tracking}

\textbf{\textsc{MultiWOZ}} \citep{budzianowski2018multiwoz,eric2019multiwoz} is a multi-domain dialogue dataset that contains 7 domains including restaurants, hotels and attractions. We use \textsc{MultiWOZ} for dialogue state tracking, the task of interpreting the user utterances throughout a dialogue in order to maintain a state representation of their requests. Dialogue state tracking is an important component of task-oriented dialogue systems; in order to fulfill a user request, it is necessary to track the user's goals over the course of multiple turns.

\section{Methods}

In this section, we describe several methods employed for the DialoGLUE tasks. We begin by describing the architectures for the 4 different tasks, all of which are built around an underlying \textsc{BERT}-like model. We then discuss \textsc{ConvBERT}, a BERT model that was fine-tuned on a large open-domain dialogue corpus. Finally, we describe our use of task-adaptive masked language modelling, which allows us to better adapt our pre-trained models to the DialoGLUE tasks.

\subsection{Architectures}

\textbf{Intent Prediction:} We fine-tune a BERT model to encode an utterance and predict its intent. Specifically, we use the pooled representation output by a BERT model and pass it through a linear layer to predict the intent class. 

\textbf{Slot Filling:} For slot filling, we represent the problem as IOB tagging \citep{ramshaw1999text} wherein every token in the utterance is labeled as either being the beginning of a slot value (B-), inside a slot value (I-) or not belonging to a slot value (O). We use BERT to produce a latent representation of each token, which is passed through a linear layer that predicts the appropriate tag (e.g., ``B-time'', ``I-people'').

\textbf{Semantic Parsing:} We transform the hierarchical representations of the \textsc{TOP} dataset into (i) a top-level intent for the utterance which corresponds to the root of the tree and (ii) a label for each word of the utterance which is the path from the root to each leaf node (which is always a word). Given this, we train a model to simultaneously predict the top-level intent for the utterance using the BERT pooled representation and the labels for each word using BERT's latent representation of each word. 

\textbf{Dialogue State Tracking:} Our state tracking architecture is inspired by the TripPy of \citet{heck2020trippy} which uses an underlying BERT model and a triple copy strategy to perform state tracking. The TripPy model uses (i) span prediction and a copy mechanism to extract values from a user utterance, (ii) a copy mechanism over concepts mentioned by the system utterance and (iii) a copy mechanism over the \textit{DS memory}, the existing dialogue state.

These architectures are held consistent throughout our experiments. Given a BERT-like encoder, we can plug it into all of the aforementioned architectures and evaluate its performance on all of the DialoGLUE tasks. This allows us to assess the quality of the underlying encoders, with confidence that performance improvements come from the improved representational power of the encoders.

\subsection{ConvBERT}

Though pre-trained models (e.g., BERT) have exhibited strong language understanding capabilities, recent work has suggested that they may be insufficient for modelling dialogue, due to dialogue's intrinsically goal-driven, linguistically diverse, multi-turn and often informal/noisy nature \citep{henderson2019convert,zhang2019dialogpt}. The unique challenges of modelling dialogue have been addressed by training on large amounts of \textit{conversational data}, from online forums. We extend these efforts by fine-tuning BERT on a large open-domain dialogue corpus consisting of nearly 700 million conversations to produce \textsc{ConvBERT}.

By training \textsc{ConvBERT} with large amounts of open-domain dialogue, we hypothesize that it is better able to produce semantically meaningful latent representations of utterances and multi-turn dialogues than a BERT model. Specifically, we fine-tune an uncased BERT-base for 4 epochs using a masked language modelling objective. Here our input representation is the last 3 turns of dialogue context followed by the [SEP] token and then the dialogue response. We truncate the entire input to have sequence length 72, and we train using the Adam optimizer with an initial learning rate of $3\cdot 10^{-4}$.

\subsection{Task-Adaptive Training}

Task-adaptive training is the process of adapting a pre-trained model to a specific task or domain, by performing self-supervised training prior to fine-tuning on the downstream task. This has been shown to help with domain adaptation \citep{mehri2019pretraining,gururangan2020don}. To adapt BERT-based encoders to the various DialoGLUE tasks, we leverage task-adaptive training. Specifically, we perform self-supervised training with the masked language modelling (MLM) objective on each dataset. We explore both (i) pre-training with MLM prior to fine-tuning on the specific task and (ii) multi-tasking by simultaneously performing self-supervised MLM and fine-tuning on the task. An example experimental setting is as follows: (1) start with the pre-trained \textsc{ConvBERT} model, (2) do MLM pre-training on the utterances of \textsc{HWU}, (3) fine-tune on \textsc{HWU} using the intent prediction architecture and simultaneously perform self-supervised MLM training on the utterances of \textsc{HWU}.

To further study the benefits of self-supervised training, we fine-tune both BERT and \textsc{ConvBERT} with masked language modelling over all the DialoGLUE datasets. In this manner we aim to adapt the two pre-trained models to task-oriented dialogue through self-supervised training.

\section{Experiments}
\begin{table*}[]
\small
\renewcommand{\arraystretch}{1.2}

    \centering
    \begin{tabular}{|l|c|c|c|c|c|c|c|c|}
    \hline
        Model   & Average & \textsc{banking77} & \textsc{hwu64}   & \textsc{clinc150}  &  \textsc{restaurant8k} & \textsc{DSTC8}   &  \textsc{TOP} & \textsc{MultiWOZ}  \\ \hline
        BERT &  86.08 & 93.02 & 89.87 & 95.93 & 95.53 & 90.05 & 81.90 & 56.30 \\
        \hspace{4px} + Pre & 86.18 & 92.34 &  91.82 & 96.27 & 95.78 & 89.48 & 81.54 & 56.07 \\
        \hspace{4px} + Multi & 85.97 &  92.27 & 90.99 & 96.22 & 95.61 & 89.93 & 81.46 & 55.30 \\
        \hspace{4px} + Pre, Multi & 85.92 & 93.20 & 90.99 & 95.67 & 95.04 & 89.96 & 82.08 & 55.06 \\
        \textsc{ConvBERT} & 86.01 & 92.95 & 90.43 & 97.07 & \textbf{95.90} & 87.58 & 82.13 & 56.00 \\
        \hspace{4px} + Pre & 86.19 & 93.25 & 92.84 & 97.09 & 95.33 & 87.02 & 82.00 & 55.67 \\
        \hspace{4px} + Multi & 85.97 & 93.20 & 91.36 & 97.09 & 95.39 & 90.02 & \textbf{82.63} & 56.48 \\
        \hspace{4px} + Pre, Multi & \textbf{86.89}  & \textbf{93.44} & 92.38 & \textbf{97.11} & 95.44 & \textbf{91.20} & 82.08 & 56.56 \\ 

        BERT-DG & 86.11 & 91.75 & 90.89 & 95.98 & 95.23 & 90.24 & 81.16 & 57.54 \\
        \hspace{4px} + Pre & 86.16 & 92.01 &  91.26 & 96.20 & 94.61 & 89.79 & 81.29 & 58.00 \\
        \hspace{4px} + Multi & 86.38 &  92.53 & 90.61 & 95.89 & 95.44 & 90.81 & 81.04 & 58.34 \\
        \hspace{4px} + Pre, Multi & 86.18 & 92.57 & 91.26 & 96.22 & 95.11 & 88.69 & 80.89 & 58.53 \\
        \textsc{ConvBERT-DG} & 82.9 & 93.21 & 91.64 & 96.96 & 93.44 & 74.54 & 72.22 & 58.57 \\
        \hspace{4px} + Pre & 84.1 & 93.05 & \textbf{92.94} & \textbf{97.11} & 95.38 & 90.88 & 60.68 & 58.65 \\
        \hspace{4px} + Multi & 82.78 & 93.02 & 91.73 & \textbf{97.13} & \textbf{95.93} & 88.97 & 53.97 & \textbf{58.70} \\
        \hspace{4px} + Pre, Multi & 85.34 & 92.99 & 91.82 & \textbf{97.11} & 94.34 & 86.49 & 76.36 & 58.29 \\

        \hline
    \end{tabular}
    \caption{Full data experiments on DialoGLUE. The average score on the DialoGLUE benchmark is shown in the leftmost column.}
    \label{tab:full}
\end{table*}
\subsection{Experimental Setup}

The experiments are carried out with four BERT-like models: \textbf{(1)} \textbf{BERT}-base,\textbf{ (2)} \textsc{\textbf{ConvBERT}} which is BERT trained on open-domain dialogues, \textbf{(3)} \textsc{\textbf{BERT-DG}} which is BERT trained on the full DialoGLUE data in a self-supervised manner with masked language modelling and \textbf{(4)} \textsc{\textbf{ConvBERT-DG}} which is \textsc{ConvBERT} trained on the full DialoGLUE data in a self-supervised manner. 

We carry out experiments with each of these four models in four different settings: (1) directly fine-tuning on the target task, (2) pre-training with MLM on the target dataset prior to fine-tuning, (3) multi-tasking with MLM on the target dataset during fine-tuning and (4) both pre-training and multi-tasking with MLM.

We perform self-supervised MLM pre-training for 3 epochs prior to fine-tuning. Fine-tuning on a target task is carried out until the performance on the validation set does not improve for 10 epochs. When multi-tasking, we alternate between fine-tuning on the target task and self-supervised training with MLM after every epoch. 

To assess the effectiveness of our pre-trained models for few-shot learning, we carry out few-shot experiments. In such experiments, self-supervised training is performed only on the few-shot data, which is 10\% of the full data. The MLM pre-training and multi-tasking is performed with only the few-shot versions of each dataset. However, both BERT-DG and \textsc{ConvBERT-DG} are trained with the full DialoGLUE data, albeit in a self-supervised manner, meaning that they see more in-domain data than either BERT or \textsc{ConvBERT} in the few-shot experiments. For all the few-shot experiments, we train five times with different random seeds and report the average performance across the five runs.
\begin{table}[h]
\renewcommand{\arraystretch}{1.2}

    \centering
\small
\begin{tabular}{|l|c|}
\hline
\multicolumn{2}{|c|}{\textsc{banking77} (accuracy)  \cmark} \\ \hline
USE \citep{casanueva2020efficient}                    & 92.81           \\
ConveRT \citep{casanueva2020efficient}                & 93.01           \\
USE + ConveRT \citep{casanueva2020efficient}            & \textbf{93.36}           \\
\textsc{ConvBERT} + Pre + Multi                 &    \textbf{93.44}                 \\ \hline

\multicolumn{2}{|c|}{\textsc{hwu64} (accuracy) \cmark} \\ \hline
USE \citep{casanueva2020efficient}                    & 91.25           \\
ConveRT \citep{casanueva2020efficient}                & 91.24           \\
USE + ConveRT \citep{casanueva2020efficient}            & 92.62           \\
\textsc{ConvBERT-DG} + Pre                 &    \textbf{92.94}                 \\ \hline

\multicolumn{2}{|c|}{\textsc{clinc150} (accuracy) \cmark} \\ \hline
USE \citep{casanueva2020efficient}                    & 95.06           \\
ConveRT \citep{casanueva2020efficient}                & \textbf{97.16}           \\
USE + ConveRT \citep{casanueva2020efficient}            & \textbf{97.16}           \\
\textsc{ConvBERT-DG} + Multi                 &    \textbf{97.13}                 \\ \hline

\multicolumn{2}{|c|}{\textsc{restaurant8k} (F-1) \cmark} \\ \hline
Span-BERT \citep{coope2020span}                    & 93.00           \\
V-CNN-CRF \citep{coope2020span}                & 94.00           \\
Span-ConveRT \citep{coope2020span}            & \textbf{96.00}           \\
\textsc{ConvBERT-DG} + Multi                 &    \textbf{95.93}                 \\ \hline

\multicolumn{2}{|c|}{\textsc{DSTC8} (F-1)} \\ \hline
Span-BERT \citep{coope2020span}                    & 91.50           \\
V-CNN-CRF \citep{coope2020span}                & 91.25           \\
Span-ConveRT \citep{coope2020span}            & \textbf{94.00}           \\
\textsc{ConvBERT} + Pre + Multi                 &    91.20                 \\ \hline

\multicolumn{2}{|c|}{\textsc{TOP} (Exact Match)} \\ \hline
RNNG \citep{gupta2018semantic}                    & 78.51           \\
SR + ELMo \citep{einolghozati2019improving}                & \textbf{87.25}           \\
\textsc{Seq2Seq-PTR}  \citep{rongali2020don}            & 86.67           \\
\textsc{ConvBERT} + Multi                 &    82.63                 \\ \hline

\multicolumn{2}{|c|}{\textsc{MultiWOZ} (Joint Goal Accuracy) \cmark} \\ \hline
DST-Picklist \citep{zhang2019find} & 53.30 \\ 
TripPy \citep{heck2020trippy}                    & 55.30           \\
SimpleTOD \citep{hosseini2020simple}                &   55.72         \\
\textsc{ConvBERT-DG} + Multi                 &    \textbf{58.70}                 \\ \hline

\end{tabular}
    \caption{Comparison to prior work on all seven datasets. We match or exceed state-of-the-art results on five out of seven datasets (marked with checkmarks), with significant improvements (\textbf{+3}) on the MultiWOZ corpus.}
    \label{tab:sota}
\end{table}
\subsection{Evaluation}

Our evaluation metrics are consistent with prior work on these datasets. For intent prediction (\textsc{banking77}, \textsc{clinc150}, \textsc{hwu64}) we use accuracy. For the slot filling tasks (\textsc{restaurant8k}, \textsc{DSTC8}), we use macro-averaged F-1 score as defined by \citet{coope2020span}. We use exact-match for \textsc{TOP}, which measures how often we exactly reconstruct the hierarchical semantic representation. For \textsc{MultiWOZ}, we use joint goal accuracy.

\begin{table*}[]
\small
\renewcommand{\arraystretch}{1.2}

    \centering
    \begin{tabular}{|l|c|c|c|c|c|c|c|c|}
    \hline
        Model   & Average & \textsc{banking77} & \textsc{hwu64}   & \textsc{clinc150}  &  \textsc{restaurant8k} & \textsc{DSTC8}   &  \textsc{TOP} & \textsc{MultiWOZ}  \\ \hline
        BERT &  66.07 & 79.87 & 81.69 & 89.52 & 87.28 & 45.05 & 74.38 & 4.69 \\
        \hspace{4px} + Pre & 66.57 & 80.72 &  83.05 & 89.73 & 86.37 & 47.17 & 74.41 & 4.55 \\
        \hspace{4px} + Multi & 66.11 &  79.89 & 82.32 & 89.69 & 87.53 & 44.92 & 74.45 & 3.95 \\
        \hspace{4px} + Pre, Multi & 66.87 & 81.49 & 82.70 & 90.53 & 86.34 & 48.55 & 74.17 & 4.29 \\
        \textsc{ConvBERT} & 68.03 & 83.63 & 83.77 & 92.10 & 86.90 & 49.08 & 74.86 & 5.90 \\
        \hspace{4px} + Pre & 67.36 & 83.68 & 83.77 & 92.10 & 86.90 & 45.20 & 74.92 & 5.09 \\
        \hspace{4px} + Multi & 68.16 & 83.15 & 82.32 & 92.33 & 86.71 & \textbf{50.49} & 75.21 & 5.48 \\
        \hspace{4px} + Pre, Multi & 68.22  & 83.99 & 84.52 & 92.75 & 86.17 & 48.40 & \textbf{78.84} & 6.87 \\ 

        BERT-DG &  72.70 & 81.47 & 83.23 & 90.57 & 85.31 & 43.85 & 74.80 &  49.70\\
        \hspace{4px} + Pre & 72.80 & 81.79 &  83.74 & 90.44 & 86.66 & 43.45 & 74.34 & 49.40 \\
        \hspace{4px} + Multi & 73.00 &  81.60 & 83.18 & 90.43 & 86.48 & 44.86 & 74.79 & 49.67 \\
        \hspace{4px} + Pre, Multi & 72.90 & 81.08 & 83.40 & 90.09 & 86.26 & 46.32 & 73.56 & \textbf{49.86} \\
        \textsc{ConvBERT-DG} & 73.75 & 84.42 & 85.17 & 92.87 & \textbf{87.65} & 41.94 & 75.27 & 48.94 \\
        \hspace{4px} + Pre & 74.10 & 84.74 & 85.63 & 93.16 & 86.95 & 43.61 & 75.32 & 49.26 \\
        \hspace{4px} + Multi & \textbf{74.35} & 84.09 & 85.74 & 93.14 & 87.48 & 45.31 & 75.37 & 49.35 \\
        \hspace{4px} + Pre, Multi & 73.80 & \textbf{85.06} & \textbf{85.69} & \textbf{93.06} & 87.58 & 44.36 & 72.01 & 48.89 \\

        \hline
    \end{tabular}
    \caption{Few-shot data experiments on DialoGLUE. The values in this table are averaged across five runs, with different random seeds.}
    \label{tab:few}
\end{table*}

\subsection{Results}

The results of the full data experiments are shown in Table \ref{tab:full}. We attain a performance gain over the vanilla BERT model \citep{devlin2018bert} across all seven datasets. These results highlight the efficacy of both the \textsc{ConvBERT} model and the task-adaptive self-supervised training. Across four datasets, the best results are attained by \textsc{ConvBERT} with both MLM pre-training and multi-tasking. We compare to prior work in Table \ref{tab:sota}, wherein we demonstrate that \textsc{ConvBERT} in combination with task-adaptive training, matches or exceeds state-of-the-art performance across five out of seven of the datasets. We attain a \textbf{+2.98} improvement in the joint goal accuracy on the dialogue state tracking task of MultiWOZ. These strong results, which hold true across several datasets, suggest that large-scale pre-training on open-domain dialogue data in combination with task-adaptive self-supervised training transfers effectively to several task-oriented dialogue tasks. 

When looking at the aggregate performance across all the DialoGLUE tasks, neither \textsc{ConvBERT} nor task-adaptive training attain improvements over BERT. However by combining these two approaches, there is a \textbf{+0.81} improvement in the average score. This suggests that through large-scale pre-training on open-domain dialogue, \textsc{ConvBERT} learns skills that are valuable to DialoGLUE, however it is only through task-adaptive training that these skills are transferred effectively to the downstream tasks. 

A noteworthy outcome of these experiments is the fact that the BERT model with task-adaptive self-supervised training sometimes outperforms \textsc{ConvBERT}. This indicates that in certain settings, it is more beneficial to perform self-supervised training on the \textit{the downstream dataset} rather than a much larger dialogue dataset. 

Performing self-supervised training across the combination of the DialoGLUE datasets gives mixed results. \textsc{ConvBERT-DG} attains a significant performance gain on MultiWOZ, suggesting that self-supervised training on other task-oriented dialogue corpora helps significantly in modelling MultiWOZ dialogues. Across other datasets, it is only marginally better than the \textsc{ConvBERT} model and sometimes worse. Aside from the unique case of MultiWOZ, it appears that self-supervised training with additional task-oriented dialogue data, beyond just the dataset in question, does not provide significant improvements. For two datasets, \textsc{DSTC8} and \textsc{TOP}, there is a decrease in performance which may be indicative of catastrophic forgetting. Namely, the \textsc{ConvBERT-DG} model may have lost the language understanding capabilities captured by the \textsc{ConvBERT} model through the additional self-supervised training, and only partially recovers this through the task-specific self-supervised training. Future work should explore better mechanisms for performing self-supervised training across the combination of the DialoGLUE datasets, as well as multi-tasking across the seven tasks. 

While our models achieve state-of-the-art performance across five of the seven tasks, they underperform on TOP and DSTC8. On the TOP dataset, the best models use sophisticated architectures which have been tailored to the task of semantic parsing \citep{einolghozati2019improving,rongali2020don}. With the DialoGLUE benchmark, our objective is to improve the underlying language encoders in a manner that results in consistent performance gains across all of the tasks. We are more concerned with the aggregate improvement across the DialoGLUE benchmark, rather than the performance on a single task. As such, we try to avoid complex task-specific architectures when simpler models achieve competitive results.

The results of the few-shot experiments are shown in Table \ref{tab:few}. The few-shot experiments are particularly important for assessing the generalizability of the methods and their ability to transfer to downstream tasks. In low data environments, self-supervised training on the entirety of the DialoGLUE datasets results in performance gains -- with BERT-DG and \textsc{ConvBERT-DG} doing better than BERT and \textsc{ConvBERT} respectively. However, this is not entirely surprising as these models are exposed to more utterances from every dataset, albeit without any of the labels. 

Most significantly, on MultiWOZ we see a 40 percent difference between \textsc{BERT-DG} and \textsc{ConvBERT-DG} over BERT and \textsc{ConvBERT}. For state tracking in particular, it appears that seeing additional dialogue data in a self-supervised setting, results in significant improvements. This may suggest that dialogue state tracking is more dependent on having semantically meaningful representations of dialogue. 

Self-supervised training on the \textit{same} dataset also helps significantly in few-shot environments. Across almost every dataset, the best result is obtained through some form of task-adaptive MLM training. Especially in settings with fewer training examples, adapting the pre-trained models to the domains of the dataset is necessary for good performance on the downstream task. 

\textsc{ConvBERT} is also far more effective in the few-shot experiments, than it was in the full data experiments with a \textbf{+1.96} point improvement in the aggregate score over BERT. While the full datasets may be sufficient to effectively transfer BERT to task-oriented dialogue, with only 10\% of the data, the benefits of the large-scale open-domain pre-training are far clearer.

\section{Conclusion}

To facilitate research into producing generalizable models of dialogue, we introduce DialoGLUE, benchmark for natural language understanding in the context of task-oriented dialogue. We experiment with several baseline methods for the DialoGLUE benchmark, demonstrating the efficacy of large-scale pre-training on open-domain dialogue and task-adaptive self-supervised training. 

To improve performance on DialoGLUE, future work should explore: (1) Large-scale pre-training that attains generalized language understanding capabilities which transfer effectively to task-oriented dialogue. (2) Mechanisms of adapting pre-trained models to specific tasks, beyond the task-adaptive masked language modelling we explore. In particular, we believe there to be potential in extending our preliminary exploration of self-supervised training on the combination of all the DialoGLUE datasets. (3) Multi-tasking across the seven datasets of DialoGLUE, as a means of transferring skills across the datasets. 

The results of our experiments demonstrate that there is significant room for improvement on DialoGLUE, particularly in the few-shot settings. The DialoGLUE benchmark is hosted publicly and we invite the research community to submit to the leaderboard.\footnote{\url{https://evalai.cloudcv.org/web/} \\ \url{challenges/challenge-page/708/}}

\bibliography{acl2020}
\bibliographystyle{acl_natbib}

\end{document}